%% file: neurips_2019.tex
\documentclass{article}

\PassOptionsToPackage{numbers, compress}{natbib}

\usepackage[final]{neurips_2019}

\input{math_commands.tex}

\usepackage[utf8]{inputenc} 
\usepackage[T1]{fontenc}    
\usepackage{hyperref}       
\usepackage{url}            
\usepackage{booktabs}       
\usepackage{amsfonts}       
\usepackage{nicefrac}       
\usepackage{microtype}      
\usepackage{tabularx}
\usepackage{graphicx}
\usepackage{subcaption}
\usepackage{amssymb}

\newcommand{\Real}[1]{\mathbb{R}^{#1}}

\title{PiNet: Attention Pooling for Graph Classification}

\author{%
  Peter Meltzer\\
  Department of Computer Science\\
  University College London\\
  London, UK \\
  \texttt{p.meltzer@cs.ucl.ac.uk} 
  \And
  Marcelo Daniel Gutierrez Mallea\\
Department of Computer Science\\
  University College London\\
  London, UK \\
  \texttt{marcelo.mallea.16@ucl.ac.uk} 
  \And
  Peter J. Bentley \\
  Department of Computer Science\\
  University College London\\
  London, UK \\
  \texttt{p.bentley@cs.ucl.ac.uk} 
}

\begin{document}

\maketitle

\begin{abstract}
  We propose PiNet, a generalised differentiable attention-based pooling
  mechanism for utilising graph convolution operations for graph level
  classification. We demonstrate high sample efficiency and superior performance
  over other graph neural networks in distinguishing isomorphic graph classes, as
  well as competitive results with state of the art methods on standard
  chemo-informatics datasets.
\end{abstract}

\section{Introduction}

Graph classification, the task of labeling each graph in a given set, has applications in many diverse domains ranging from chemo-informatics \cite{DeCao2018} and bio-informatics \cite{Zitnik}, to image classification \cite{Harchaoui} and cyber-security \cite{Chau2011}. In recent years, Convolutional Neural Networks (CNNs) have led the state of the art in many forms of pattern recognition, i.e. in images \cite{Graham2014} and audio \cite{Aytar2016}.

Essential to the success of CNNs in representation learning is the process
of pooling \cite{LeCun:1998:CNI:303568.303704}, in which a set of related
vectors are reduced to a single vector (or smaller set of vectors). An important
property of a pooling operator is invariance to different orderings of the input
vectors. In vertex level learning tasks such as link prediction and vertex
classification, Graph Convolutional Networks (GCNs) achieve invariance by
pooling neighbours' feature vectors with symmetric operators such as
feature-weighted mean \cite{Kipf2016}, max \cite{Hamilton2017c}, and
self-attention weighted means \cite{Velickovivelickovic2018}.

In this work we present PiNet\footnote{Code available at
\url{http://github.com/meltzerpete/pinet}},
a differentiable pooling mechanism by which the vertex-level invariance to
permutation achieved for vertex level tasks may be extended to the graph level.
Inspired by the attention mechanisms of RNNs \cite{Mnih2014} and Graph
Attention Networks (GAT) \cite{Velickovivelickovic2018}, we propose an
attention-based aggregation method which weights the importance of each vertex
in the final representation.

\section{Related work}

The idea of permutation invariant deep learning is not new. \cite{Zaheer2017}
consider the case of classification on sets, in which they propose that a
permutation invariant function $f(\sX)$ on the set $\sX$ may be learned indirectly
through decomposition in the form
\begin{equation}
  f(\sX) = \rho \left( \sum_{x \in \sX} \phi(x) \right),
  \label{eq:deepsets}
\end{equation}
if suitable transformations $\rho$ and $\phi$ can be found. This idea is
specialized as Janossy Pooling in \cite{Murphy2019}, where $\rho$
is a normalisation function, and the summation occurs over the set of all
possible permutations of the input set. They also propose the
use of canonical input orderings and permutation sampling offering a trade-off
between learnability and computational tractability.

The use of canonical orderings to tackle permutations in graph representation
learning has been demonstrated to be effective in Patchy-SAN \cite{Niepert2016}.
Here canonical labellings are applied to provide an ordering over which nodes
are sampled, aggregated and normalised to convert each graph to a fixed sized
tensor which is then fed into a traditional CNN. DGCNN \cite{Zhang} also uses a
sorting method to introduce permutation invariance, where vertex embeddings are
first obtained with a GCN, and then sorted before being fed into a traditional CNN.

Considering the task of vertex classification, the GCN as introduced by
\cite{Kipf2016} can in fact be formulated as a particular instance of
\autoref{eq:deepsets}, where for each vertex $i$ the output $\vx_{i}^{(l+1)}$ of
a single layer $l$ with input features $\vx^{(l)}$ is given by
\begin{equation}
  \label{eq:gcn}
  \vx_{i}^{(l+1)}
  = \sigma \left(
    \sum_{j \in \mathcal{N}(i) \cup \{i\}} \frac{1}{c_{ij}} \vx_{j}^{(l)} \mW^{(l)}
  \right),
\end{equation}
where $\sigma$ is a non-linear activation function, $\mathcal{N}(i)$ is set of
vertices in the immediate neighbourhood of vertex $i$,
$c_{ij}$ is a normalisation constant of edge $(i, j)$, and $\mW^{(l)}$ is the
learned weights matrix for layer $l$. We also note that \cite{Kipf2016} and
variants may also be expressed as an instance of the Weisfeiler-Lehmen graph
isomorphism algorithm \cite{Weisfeiler1968}, thus providing the theoretical justification for which
graph convolution operations are able to capture the structural information of
graphs.

\cite{Velickovivelickovic2018} extends
\cite{Kipf2016} with the introduction of attention mechanisms, where a vertex's
edges are weighted by a neural network with the vertex pair as input. Many other
(in fact virtually all) variants of \cite{Kipf2016}, i.e.
\cite{Defferrard2016,Hamilton2017c,Velickovivelickovic2018,Morris2019,Xu,Wu2019},
may also be expressed as an instance of \autoref{eq:deepsets}, therefore
indicating invariance to permutations at the vertex level (GraphSAGE with LSTM
neighbourhood aggregator \cite{Hamilton2017c} is an example of one that is not).
However, since the vertices have no natural ordering, the output matrix of a GCN
is not inherently invariant to permutation and thus
does not make a good graph representation.

A simple solution is to use a symmetric operator to combine vertex vectors to
form a single graph vector, for example the mean. Again we can formulate this
entire process as an instance of \autoref{eq:deepsets}, where $\rho$ is the
mean, and $\phi$ is the GCN's particular vertex function. A less naive method to
aggregate GCN-learned vertex embeddings can be seen in DiffPool
\cite{Ying2018b}, where GCN-based vertex embeddings are used to cluster nodes to
aggregate features hierarchically, thus considering the structural information of
the graph as opposed to a flat, global aggregation. Other structural pooling
methods include \cite{Lee2018a} which use attention-based guided walks to direct
RNNs to select parts of the graph to inform the final representation.

\section{PiNet}

\subsection{Model architecture}

PiNet is a generalised end-to-end deep neural network architecture that utilizes the
vertex-level permutation invariance of graph convolutions in order to learn graph
representations that are also invariant to permutation.

\begin{figure}[htb]
  \centering
  \includegraphics[width=.8\linewidth]{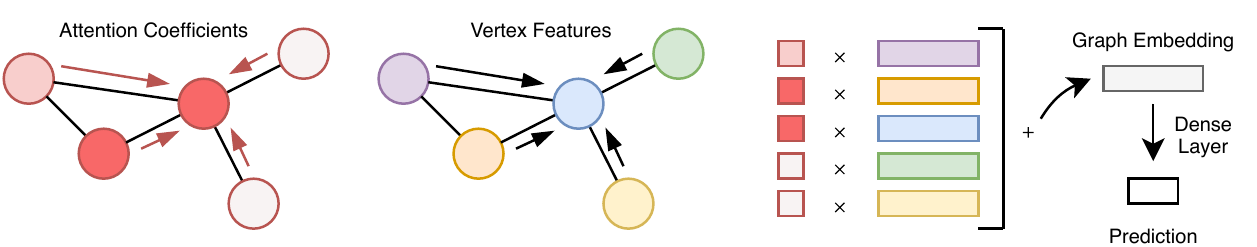}
  \caption{Overview of PiNet: One message passing network learns vertex
  features, the other learns attention coefficients. The final graph
representation is a sum of the learned vertex features weighted by the attention
coefficients. For multiple attention dimensions per vertex, the graph embedding
becomes a matrix where the rows are concatenated to form a single vector.}
  \label{fig:architecture}
\end{figure}

Let $G = (\mA, \mX)$ be a graph from a set $\mathcal{G}$ with adjacency matrix
$\mA \in
\Real{N \times N}$ and vertex features matrix $\mX \in \Real{N \times F}$,
and $\psi: (\Real{N \times N}, \Real{N \times F}) \rightarrow
\Real{N \times F'}$ be any message passing convolution network (i.e.
the GCN \cite{Kipf2016}) (note $\psi$ may contain an arbitrary number of layers).
PiNet may then be defined by the output for a single graph,
\begin{equation}
  \label{eq:z}
  z(G) = \sigma_S \Bigl[ g\Bigl( \sigma_S \bigl(\left[\psi_A\left(\mA,
  \mX\right) \right]^\top\bigr) 
  \cdot \psi_X\left(\mA, \mX\right) \Bigr) \mW_D \Bigr] \in \Real{C},
\end{equation}
where $\sigma_s$ is the softmax activation function, $g$ is a function that
concatenates rows of a matrix to form a vector, $\psi_A$ and
$\psi_X$ are separate message passing networks for learning attention
coefficients and vertex features respectively, $\cdot$ is a matrix product, $\mW_D$ is a weights
matrix for a fully connected dense layer, and $C$ is the number of target
classes. The inner softmax constrains the attention coefficients to sum to 1 and
prevents them from all falling to 0. The outer softmax may be replaced 
for multi-label classification tasks (i.e. sigmoid).

\section{Experiments}

All hyper-parameters are detailed in \autoref{app:hypers}.

\subsection{Datasets}

For the isomorphism test (\ref{se:isomorphism}) we use a generated dataset
available from our repository. The generation process is detailed in
\autoref{app:data}. All other experiments are performed using a
standard set of chemo-informatic benchmark datasets\footnote{Available at
\url{https://ls11-www.cs.tu-dortmund.de/staff/morris/graphkerneldatasets}.}.

\subsection{Experiment 1: Isomorphism test}
\label{se:isomorphism}

For PiNet we use $\psi_A = \psi_X = \sigma_R ( \tilde{\mA} \cdot \,
  \sigma_R ( \tilde{\mA} \,\, \mX \mW^{(0)} ) \,\, \mW^{(1)} )$ \cite{Kipf2016},
where $\tilde{\mA} = \mD^{-\frac{1}{2}} \hat{\mA} \mD^{-\frac{1}{2}}$, $\mD$ is the diagonal degree matrix of
$\hat{\mA}$, $\hat{\mA} = \mA + \mI$,
$\mI$ is the identity matrix, and
$\sigma_R$ is the ReLU activation function. We refer to this as PiNet (GCN). To evaluate the performance of our
proposed architecture directly, we compare against a GCN with a dense layer
applied to the concatenated vertex vectors and a GCN with a dense layer on the
mean of its vertex vectors.

We also compare with three state of the art graph classifiers:
DiffPool \cite{Ying2018b}, DGCNN \cite{Zhang}, and Patchy-SAN \cite{Niepert2016}. We vary the number of training examples using stratified sampling and report the
mean validation accuracy of 10 trials.

\subsection{Experiment 2: Message passing mechanism}
\label{se:message}

We extend the message passing matrix of \cite{Kipf2016} in which we add two
additional trainable parameters, thus vector state is propagated by the matrix
\begin{equation}
  \tilde{\mA} = (p \mI + (1 - p) \mD)^{-\frac{1}{2}} (\mA + q \mI) (p \mI + (1 - p) \mD)^{-\frac{1}{2}},
  \label{eq:messages}
\end{equation}
where $\mI$ is the identity matrix, $\mD$ is the diagonal degree matrix,
and $\mA$ is the graph adjacency matrix. $p$ allows the model to optimise
the extent to which to apply symmetric normalisation of the adjacency matrix,
and $q$ (as originally supposed for further work in \cite{Kipf2016}) allows the model to optimise
the trade-off between keeping a vertex's own state and aggregating the states of
its neighbours. Note that $p$ and $q$ are learned indirectly through optimising
$p'$ and $q'$ with sigmoid to give $0 \le p, q \le 1$.

We compare the classification accuracy of the extreme cases of $p$ and $q$
($\mA$, $\mA + \mI$, $\mD^{-\frac{1}{2}} \mA
  \mD^{-\frac{1}{2}}$, and $\mD^{-\frac{1}{2}} (\mA + \mI)
\mD^{-\frac{1}{2}}$)
 against the learned $p$ and $q$ for each layer in each
head. Following the methodology of \cite{Xu} and
\cite{Ying2018b} we perform 10-fold cross validation, reporting the mean
validation accuracy for the single best epoch across the folds.

\subsection{Experiment 3: Benchmark}

We benchmark the performance of PiNet with the original \cite{Kipf2016} and
extended GCNs (\ref{se:message}), on the benchmark datasets described above
against the baseline and state of art methods used in \ref{se:isomorphism},
using the methodology as described in \ref{se:isomorphism}. 

\section{Results}

\begin{figure}[h]
\centering
\begin{minipage}{.38\textwidth}
        \centering
	\includegraphics[width=\textwidth]{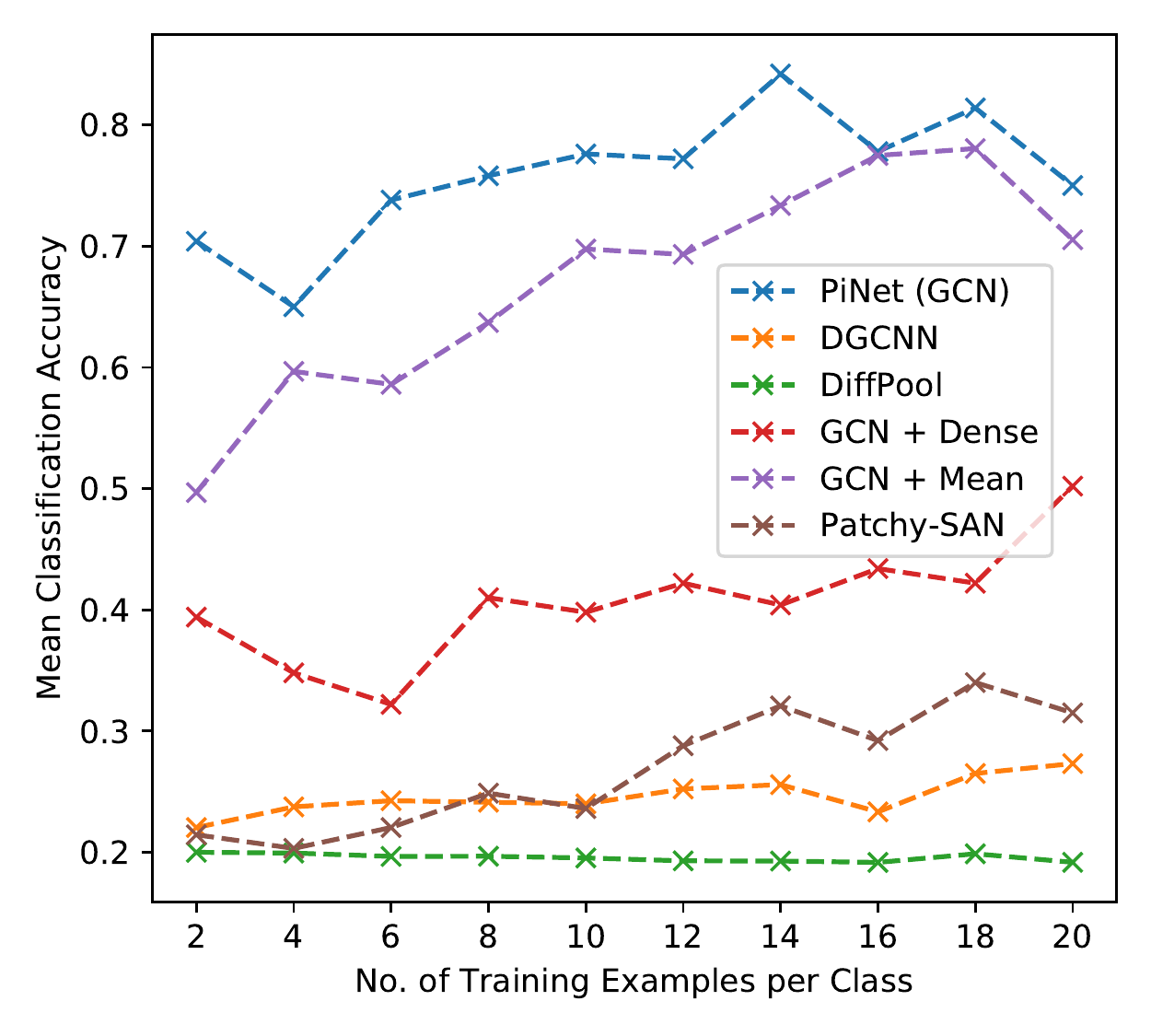}
\end{minipage}\quad%
\begin{minipage}{.5\textwidth}
        \centering
	\includegraphics[width=\textwidth]{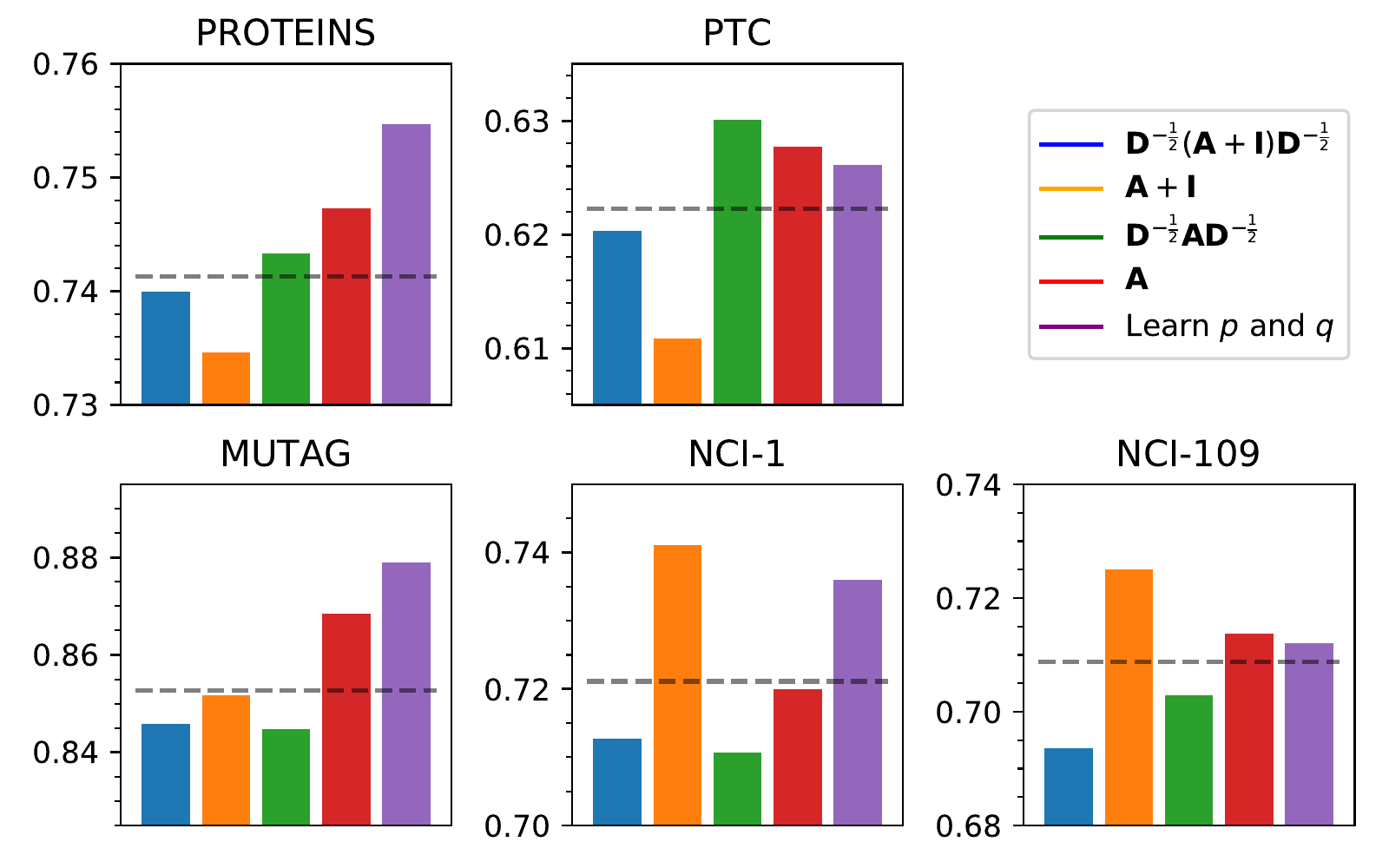}
\end{minipage}
\begin{minipage}[t]{.38\textwidth}
        \centering
	\caption{Mean classification accuracy over a range of training set sizes
	on the isomorphism dataset.}
  \label{fig:isomorphism}
	\vfil
\end{minipage}\quad%
\begin{minipage}[t]{.5\textwidth}
        \centering
  \caption{Mean classification accuracy for each message passing matrix within
  PiNet. Dashed lines indicate mean accuracy of manual search.}
  \label{fig:all_matrices}  
\end{minipage}
\end{figure}

\begin{table}[h]
  \centering
  \caption{Benchmark results. * indicates GCN with extended message passing with
    manual $p$ and $q$, and ** with learned $p$ and $q$.}
  \begin{tabularx}{\linewidth}{Xlllll}
  \toprule
   &            MUTAG &            NCI-1 &          NCI-109 &         PROTEINS &              PTC \\
  \midrule
  GCN + Dense &  $0.86 \pm 0.06$ &  $0.73 \pm 0.03$ &  $0.72 \pm 0.02$ &  $0.71 \pm 0.04$ &  $0.63 \pm 0.07$ \\
  GCN + Mean  &  $0.84 \pm 0.07$ &  $0.68 \pm 0.03$ &  $0.67 \pm 0.03$ &  $0.74 \pm 0.02$ &  $0.63 \pm 0.04$ \\
  Patchy-SAN  &  $0.85 \pm 0.06$ &  $0.58 \pm 0.02$ &  $0.58 \pm 0.03$ &  $0.70 \pm 0.02$ &  $0.58 \pm 0.02$ \\
  DGCNN       &  $0.86 \pm 0.07$ &  $0.73 \pm 0.03$ &  $0.72 \pm 0.02$ &  $0.73 \pm 0.05$ &  $0.61 \pm 0.06$ \\
  DiffPool    &  $0.91 \pm 0.08$ &  $0.73 \pm 0.02$ &  $0.72 \pm 0.03$ &  $0.80 \pm 0.05$ &  $0.64 \pm 0.07$ \\
  \midrule
  PiNet (GCN)  &  $0.85 \pm 0.07$ &  $0.71 \pm 0.03$ &  $0.69 \pm 0.03$ &  $0.74
  \pm 0.05$ &  $0.62 \pm 0.05$ \\
  PiNet (GCN*) &  $0.87 \pm 0.08$ &  $0.74 \pm 0.03$ &  $0.73 \pm 0.03$ &  $0.75 \pm 0.06$ &  $0.63 \pm 0.06$ \\
  PiNet (GCN**)&  $0.88 \pm 0.07$ &  $0.74 \pm 0.02$ &  $0.71 \pm 0.04$ &  $0.75 \pm 0.06$ &  $0.63 \pm 0.04$ \\
  \bottomrule
  \end{tabularx}
\end{table}

Experiment 1 (\autoref{fig:isomorphism}) demonstrates the power of PiNet
in capturing the most subtle differences between the test graphs, even with
only 2 examples per class. Interestingly, this data presents a
worst-case scenario for DiffPool and thus this method is unable to distinguish the different
graph classes at all. In Experiment 2
(\autoref{fig:all_matrices}) we see that while the optimal parameters $p$ and
$q$ are not always found, the result of learning $p$ and $q$ offers better
performance than the average of a manual search over the extreme values in all
cases thus suggesting it is a suitable technique to reduce parameter searching.
Finally, for the standard benchmark datasets we observe competitive performance
with (within one standard deviation or better than) the state of the art methods for all
datasets.

\section{Conclusion}

We have introduced PiNet, a generalised attention-based pooling mechanism for utilizing vertex-level
convolution operators for graph level representations. We have demonstrated its
ability to capture the finest subtleties in a graph isomorphism test and
demonstrated results competitive with current state of the art methods on
standard benchmark datasets. For further work we propose further study of PiNet
with different convolution operators, as well as the use of skip connections to
add great flexibility to the learned vertex representations prior to graph level
pooling.

\subsubsection*{Acknowledgments}

We thank Braintree Ltd. (\url{http://braintree.com}) for providing the full
funding for this research.

\bibliography{neurips_2019}

\appendix

\section{Hyper-Parameters}
\label{app:hypers}

In all experiments we use categorical cross-entropy for loss,
and fix learning rate to $10^{-3}$.

\begin{itemize}
  \item PiNet (GCN): hidden sizes $\{32, 64\}$ for each layer in each head (two
    layers).
  \item GCN + Dense \& GCN + Mean: hidden sizes $\{32, 64\}$ for each layer (two
    layers).
  \item DiffPool: assign-ratio in $\{0.1, 0.2, 0.3\}$, hidden layer sizes in $\{30, 40, 50\}$ (for two layers)
  \item DGCNN: hidden sizes in $\{64, 96, 128\}$ and 3 sort pooling values selected
according to the size of each dataset.
  \item Patchy-SAN: labelling procedures: NAUTY \citep{Mckay2014} and Betweenness Centrality \citep{Brandes2001}.
\end{itemize}

\section{Isomorphism Dataset Generation}
\label{app:data}

To generate the
data we sample 5 unique Erd\~{o}s-R\'{e}nyi graphs \cite{Erdos1960} with 
equal vertex degree distributions - this ensures a high level of challenge and prevents
trivial classification. Each vertex is assigned one of two classes uniform
randomly. The 5 unique graphs are then copied
99 times each and the vertex ids are permuted randomly on all of the
graphs since we wish to test the ability to recognise isomorphic graphs even
with different vertex orderings. 

\end{document}

%% file: math_commands.tex
\usepackage{amsmath,amsfonts,bm}

\def\eqref#1{equation~\ref{#1}}

\def\1{\bm{1}}

\def\vx{{\bm{x}}}

\def\mA{{\bm{A}}}

\def\mD{{\bm{D}}}

\def\mI{{\bm{I}}}

\def\mW{{\bm{W}}}
\def\mX{{\bm{X}}}

\DeclareMathAlphabet{\mathsfit}{\encodingdefault}{\sfdefault}{m}{sl}
\SetMathAlphabet{\mathsfit}{bold}{\encodingdefault}{\sfdefault}{bx}{n}

\def\sX{{\mathbb{X}}}

